\documentclass[letterpaper, 10 pt, conference]{ieeeconf}

\usepackage[left=54pt,top=50pt,right=54pt,bottom=57pt]{geometry}

\IEEEoverridecommandlockouts  % for \thanks
\usepackage[utf8]{inputenc}
\usepackage{moreverb,url}
\usepackage{times}
\usepackage{multicol}
\usepackage{amsmath,amsfonts,amssymb,mathtools,leftidx}%,amsthm}
\usepackage{wrapfig}
\usepackage{float}
\usepackage{xcolor}
\usepackage{graphicx}
\usepackage{multirow}
\usepackage{mathrsfs}
\usepackage{nccmath}
\usepackage{balance}
\usepackage{afterpage}

\usepackage{tabularx}
\usepackage[noadjust]{cite}
\usepackage[colorlinks,bookmarksopen,bookmarksnumbered,citecolor=red,urlcolor=red]{hyperref}
\usepackage{booktabs}
\usepackage[ruled,vlined,linesnumbered,algo2e]{algorithm2e} % for algorithm

\usepackage{algorithm,algpseudocode}

\makeatletter
\newcommand\fs@spaceruled{\def\@fs@cfont{\bfseries}\let\@fs@capt\floatc@ruled
  \def\@fs@pre{\vspace{8pt}\par\noindent\rule[2pt]{\columnwidth}{1pt}}%height.8pt depth0pt
  \def\@fs@post{\vspace{1pt}\par\noindent\rule[0pt]{\columnwidth}{1pt}}%
  \def\@fs@mid{\vspace{0pt}\par\noindent\rule[2pt]{\columnwidth}{1pt}}%
  \let\@fs@iftopcapt\iftrue}
\makeatother

\begin{document}

\title{\LARGE \bf
Enhancement for Robustness of Koopman Operator-based \\Data-driven Mobile Robotic Systems}
\author{Lu Shi and Konstantinos Karydis
\thanks{The authors are with the Dept. of Electrical and Computer Engineering, University of California, Riverside. 
	Email: {\{lshi024, karydis\}@ucr.edu}.}
\thanks{We gratefully acknowledge the support of NSF under grant \# IIS-1910087 and of ONR under grant \# N00014-19-1-2264. Any opinions, findings, and conclusions or recommendations expressed in this material are those of the authors and do not necessarily reflect the views of the funding agencies.}% <-this % stops a space
}

\maketitle
\thispagestyle{empty}

\begin{abstract}
Koopman operator theory has served as the basis to extract dynamics for nonlinear system modeling and control across settings, including non-holonomic mobile robot control. There is a growing interest in research to derive robustness (and/or safety) guarantees for systems the dynamics of which are extracted via the Koopman operator. In this paper, we propose a way to quantify the prediction error because of noisy measurements when the Koopman operator is approximated via Extended Dynamic Mode Decomposition. We further develop an enhanced robot control strategy to endow robustness to a class of data-driven (robotic) systems that rely on Koopman operator theory, and we show how part of the strategy can happen offline in an effort to make our algorithm capable of real-time implementation. We perform a parametric study to evaluate the (theoretical) performance of the algorithm using a Van der Pol oscillator, and conduct a series of simulated experiments in Gazebo using a non-holonomic wheeled robot.
%at run-time use the estimated prediction error for  to offer robustness%The formulation could be further utilized in control, planning and navigation as a fixing term. 
%Also, to enable the algorithm in a physical robot considering computation time, space and energy, we provide a comprehensive procedure that extract the parameters which could be prepared offline %as illustrated in Fig.~\ref{fig:algorithm}. 
%A simulation study with a nonlinear oscillator validates theoretical results and we test the approach in a ROSbot 2.0 robot.
\end{abstract}

%%%%
%%%%
\section{Introduction}

%4: Koopman operator
%5: Limitation on Koopman operator approaches - prediction error
%6: What do we propose
%7: Key contributions
%i) prediction error formulation, ii) algorithm to deploy an adaptive structure for robot navigation, iii) testing with hardware.

Employing models for mobile robot motion planning and control can be beneficial for integrating motion constraints in planning~\cite{LavalleBook,sharma2017mpoverview} and deriving control performance guarantees (e.g., robustness guarantees~\cite{zheng2020safeMobile,da2016safeguardmb}). 
Yet, there exist many instances in which robots interact physically with their environment and that these interactions are uncertain. Examples include robots operating in partially-known dynamic environments~\cite{aoude_AURO_13}; legged robots traversing non-smooth terrains~\cite{LDog_IJRR_11,qianRSS}; quadrotors flying under the influence of uncertain aerodynamic effects~\cite{gusts_AIAA_10, caitlin,RAL,CCTA}; underwater robots affected by uncertain ocean currents~\cite{AUV-jfr-13}; and steerable needles interacting with soft tissue~\cite{alterovitz-IJRR-08}. 
Thus, employing pre-selected models may restrict the capability to predict actual robot behaviors when operating under uncertainty~\cite{Tim_ICRA_93, Thrun_2005,Chiri_book_2012b, IJRR}.

A different way to address uncertainties in robot-environment interactions is by extracting dynamics from data. %~\cite{karydis2016kPOD,moreREFS}. 
Multiple distinct approaches have been proposed---including stochastic model extension~\cite{IJRR}, (deep) reinforcement learning~\cite{abdulsamad2020rl}, (deep) neural networks~\cite{sindhwani2020ml, schmidhuber2015DNN_book}, and spectral methods~\cite{kaiser2019overview}, to name a few---and deployed in robotics (e.g.,~\cite{IJRR,suprijono2017NNR,Shi2019neuralLander,shi2020neuralswarm,taylor2019NNR,CCTA}). 
The advent of data-driven methods for modeling complex dynamical systems creates a need for theoretical safety and/or performance guarantees. 

Deriving guarantees for methods that rely on extracting dynamics from data for mobile robots is a rapidly-growing focus area~\cite{cui2020guarantee}. Several efforts focus on cases that involve deep neural networks, e.g., by developing spectrally-normalized margin bounds~\cite{Bartlett2017margin} or by embedding and/or extending Lyapunov stability theory in neural network-based systems for safety (e.g.,~\cite{Aswani2013LearningMPC,Berkenkamp2017RLstability,fisac2018safetylearning,berkenkamp2015safelearning,akametalu2014reachability}). % 
Despite their high expressiveness, neural network-based systems continue to require large training data sets, and might lead to instability and unpredictable outputs because of overfitting. 
%
%While as robots increasingly venture outside of the lab, the derived models often have limited use or poor prediction over longer time spans under the uncertainty generated by interacting with environment or other agents. Recently, people turn to use data-driven methods for model extractions. The advent of data-based methods for modeling and control of complex dynamical systems creates a need to come up with theoretical safety and/or performance guarantees. 

%One of the data-driven based researches is integrating various machine learning approaches, such as Gaussian processes~\cite{rasmussen2003GP_book} or deep neural networks~\cite{schmidhuber2015DNN_book}, within robotics (e.g.,~\cite{suprijono2017NNR,Shi2019neuralLander,shi2020neuralswarm,taylor2019NNR}). Due to their high expressiveness, deep neural networks, in particular are currently at the forefront of research on stability of data-driven control system methods. 
%
%For instance,~\cite{Bartlett2017margin} present spectrally-normalized margin bounds for methods (~\cite{Shi2019neuralLander}) using ML approach to build black-box models. Another approach is to embed Lyapunov stability theory in NN-based systems for safety (e.g.,~\cite{Aswani2013LearningMPC,Berkenkamp2017RLstability,fisac2018safetylearning,berkenkamp2015safelearning,akametalu2014reachability}). While learning algorithms with NNs can achieve impressive results, they continue to require large training data sets, and might lead to instability and unpredictable outputs because of overfitting.
%%

%%%
Besides (deep) neural networks, dimensionality reduction and spectral approaches play an important role in data-driven algorithms~\cite{kaiser2019overview}. Methods like Proper Orthogonal Decomposition~\cite{chatterjee2000POD,karydis2016kPOD}), Dynamic Mode Decomposition (DMD)~\cite{Tu2013DMD} and Extended DMD (EDMD)~\cite{williams2015EDMD}, and their various extensions~\cite{klus2020koopmangenerator,kaiser2017koopmaneigenfunction,folkestad2019koopmaneigenfunction,brunton2016koopmanInvariantSpace} have been successfully applied across areas. Most of the techniques turn out to be strongly related to the Koopman operator theory~\cite{koopman1931koopman}, which is a powerful tool to extract complex dynamics from data. 

The Koopman operator theory can be used to map a finite-dimensional nonlinear system to an infinite-dimensional linear system that evolves by a linear operator. %The theory was first proposed by. 
The Koopman operator has been used for model-based control of dynamical systems, including feedback stabilization~\cite{huang2019feedback}, optimal control~\cite{abraham2019optimalControl,abraham2017optimalControl}, and model predictive control~\cite{bruder2019mpc,arbabi2018mpc,korda2018KoopmanMPC}.
Recent research efforts have also sought to use the Koopman operator in robotics, such as in robot control~\cite{todd2020derivative,abraham2017optimalControl}, modeling of soft robots~\cite{bruder2019softK,castano2020softK}, and human-machine interaction~\cite{broad2020humanmachine}.  
There also exist some works to derive guarantees for methods employing the Koopman operator, including investigation of convergence of estimation~\cite{korda2018convergence,korda2018KoopmanMPC,peitz2019switchsystem} and global error bounds for the operator~\cite{todd2020derivative,mamakoukas2020stable}. However, investigating the 
prediction error of, or providing robustness guarantees for, the perturbed systems' performance when the data used for modeling are noisy remains under-developed. The present paper addresses this gap.

Contrary to traditional robust control~\cite{zhou1998robustControl}, noisy observations disturb not only the controller itself but also the data used to estimate a model for the system (and which is used for control). Hence, there is need for new tools that can capture the intricacies of concurrent data-driven analysis and control, and which are fundamentally different from traditional robust control theory. 

\begin{figure}[t!]
\vspace{6pt}
\centering
\includegraphics[width = 0.48\textwidth]{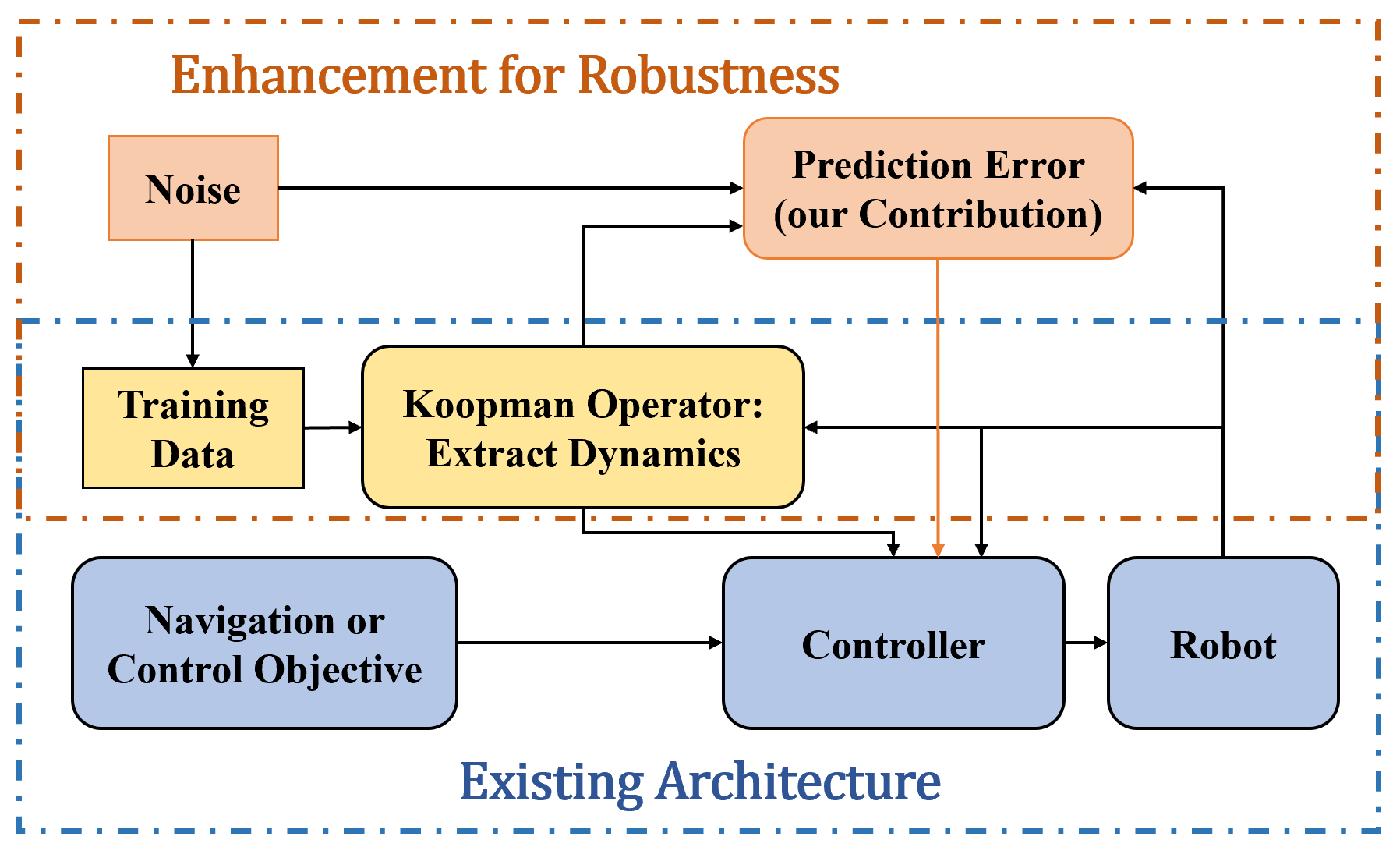}
\vspace{-6pt}
\caption{Overview of the enhanced robot control structure proposed in this work. }
\label{fig:system}
\vspace{0pt}
\end{figure}

In this paper, we first develop an approach to quantify the prediction error because of noisy data when approximating a model for control of a data-driven system via the Koopman operator. To estimate the Koopman operator we employ a variant of EDMD with control (EDMDc)~\cite{proctor2018KoopmanInputs} which generalizes EDMD to forced systems. The proposed formulation relies on studying the sensitivity of components of the Koopman operator to noisy data. (Table~\ref{tab:notations} contains some key notation used in this paper.) Further, we propose an algorithm for an enhanced robot control structure that endows robustness to the data-driven system at hand. The proposed algorithm is designed to work in unison with an existing data-driven robot control architecture to add robustness without replacing parts of the underlying architecture (Fig.~\ref{fig:system}). 
We address practical aspects on how to deploy the proposed algorithm across nonlinear dynamical systems and mobile robots, and show how to make (parts of) the algorithm run online to enable real-time implementation.  
Lastly yet importantly, the proposed approach may generalize across data-driven systems. We show evidence of generalizability via a parametric simulation study using a Van der Pol oscillator and experimentation in Gazebo with a non-holonomic wheeled robot (ROSbot2.0) under distinct noise levels. 

Succinctly, the main contributions of this work include:
\begin{itemize}
    \item Quantification of prediction error caused by noisy data for control of data-driven systems that employ the Koopman operator to approximate the systems' model.
    \item Development of an algorithm to endow robustness to data-driven systems that employ the Koopman operator.
    \item Evaluation of the method's efficacy and generalizability via testing with a nonlinear dynamical system and with a non-holonomic wheeled robot.
\end{itemize}
%

%Succinctly, the main contributions of this paper is to derive the prediction error based on the model generated by Koopman operator using EDMDc under noisy training data. We take care of details when implement it in physical robots and provide an exhaustive procedure. The efficiency of our work is firstly shown in a Van del Pol oscillator system within different setting of parameters. Finally we implement the algorithm in a physical autonomous robot platform, ROSbot 2.0 robot under different setting of noise.

%After presenting briefly necessary technical preliminaries (Section~\ref{sec:background}), we formulate the prediction error expression (Section~\ref{sec:prediction}) and subsequently show the stability of the derived stochastic system (Section~\ref{sec:stability}). A simulation study evaluates the theoretical analysis (Section~\ref{sec:simulation}). Along with summarizing key findings, we also offer insight on the effect of the number of measurements used for online update of approximating the Koopman operator (Section~\ref{sec:discussion}).
%%%%

\vspace{-3pt}
\begin{table}[h!]
\vspace{-4pt}
\caption{List of key notation used in the paper.}
\label{tab:notations}
\vspace{-12pt}
\begin{center}
\begin{tabular}{l|l}
\hline
Notation & Description  \\
\hline
$N_x$& Dimension of state $x$ \\
\hline
$N_u$& Dimension of input $u$  \\
\hline
$t$& Index for online system propagation  \\
\hline
$\mathcal{K}$ & Koopman operator\\
$K$ & Approximated Koopman operator\\
\hline
$v_q$& $q$-th Koopman mode\\

$\lambda _q$& $q$-th Koopman eigenvalue\\ %(eigenvalue of Koopman operator) \\

$\varphi_q$ & $q$-th Koopman eigenfunction\\
\hline
\multirow{2}{*}{$Q$}& Dimension of observables' dictionary;\\
& Size of estimated Koopman operator \\
\hline
$\xi_q$ & $q$-th right eigenvector of Koopman operator\\
$w_q$& $q$-th left eigenvector of Koopman operator\\
\hline
$\mathbf{\Psi}$ & Vector-valued observation\\
\hline
\end{tabular}
\end{center}
\vspace{-14pt}
\end{table}

%%%%%%%%%
\section{Preliminary Technical Background on Koopman Operator Theory and EDMD}\label{sec:background}

%%%%%%%%%%%%%%%%%%%%%%%%%%%%%%%%%%%%%%%%%%%%

The Koopman operator is an infinite-dimensional linear operator that governs the evolution of observables $g(x_t,u_t)$ of the original states. The evolving operator $f$ of the original system can be represented by Koopman modes, eigenvalues, and eigenfunctions.  In this section, we give an overview of key relevant results on Koopman operator theory and EDMD on how to extract system dynamics from data.

In its original formulation, Koopman operator theory applies to unforced systems. There are two ways to generalize to forced systems. The first is to lift states and inputs into two spaces separately, and then design controllers for the lifted system~\cite{williams2016KoopmanInputs}. The other way is to extend the Koopman operator theory with control for systems with nonlinear input-output characteristics~\cite{proctor2018KoopmanInputs}. We adopt the second approach.\footnote{We consider the most general formulation in which inputs are generated from an exogenous forcing term. For details on other formulations the reader is referred to~\cite[Section 3.1]{proctor2018KoopmanInputs}.} 

Consider the forced nonlinear dynamical system 
\begin{equation}\label{eq:DSystem}
    x_{t+1} = f(x_t, u_t) \enspace,
\end{equation}
where $ x\in \mathbb{R}^{N_x}$ and $u \in \mathbb{R}^{N_u}$. %with a set of snapshots {\small $X = [x_1,x_2,\dots,x_{M},x_{M+1}]$}, and $U = [u_1,u_2,\dots,u_{M},u_{M+1}]$. 
Define a set of observables that are functions of both the states and inputs where $g:\mathbb{R}^{N_x}\times \mathbb{R}^{N_u} \to \mathbb{R}^{N_x+N_u}$. The propagation law of observables $g$ with the Koopman operator is
$\mathcal{K} g(x_t, u_t)= g(f(x_t, u_t),u_{t+1})$. Then, under the full-observability assumption such that $g(x_{t+1},u_{t+1})=[x_{t+1}; u_{t+1}]$ and decomposing with $Q$ Koopman modes $v_q$, eigenvalues $\lambda _q$ and eigenfunctions $\varphi_q$, we obtain as in~\cite{williams2015EDMD}:
$$[x_{t+1}; u_{t+1}] = g(f(x_t, u_t),u_{t+1})= \mathcal{K} g(x_t, u_t)$$
\begin{equation}\label{eq:extendPredicion}
   \to [x_{t+1}; u_{t+1}]  =\sum_{q=1}^Q v_q\lambda_q\varphi_q(x_t, u_t)\enspace . 
\end{equation}
The predicted state $x_{t+1}$ is obtained by taking the first $N_x$ elements of the vector in the right-hand part of~\eqref{eq:extendPredicion}.

Consider sets (termed snapshots) of $M+1$ state measurements and $M+1$ associated control inputs, that is {\small $X = [x_1,x_2,\dots,x_{M},x_{M+1}]$}, and $U = [u_1,u_2,\dots,u_{M},u_{M+1}]$. 
A way to estimate the Koopman operator $\mathcal{K}$ from $M+1$ state and control snapshots is via Extended Dynamic Mode Decomposition (EDMD). EDMD generates a finite dimensional approximation $K$ of $\mathcal{K}$. It does so by employing a dictionary of functions to lift state variables to a space where dynamics is approximately linear. 

Given a dictionary of observables of size $Q$, $\mathcal{D} = \left\{\psi_{1}, \psi_{2}, \ldots, \psi_{Q}\right\}$, where each dictionary element $\psi_q$, $q=1,\ldots,Q$ is a differentiable function containing $x_m$ and $u_m$ terms, set the vector-valued dictionary as $\mathbf{\Psi} = [\psi_1, \dots, \psi_Q]$.\,\footnote{Examples of dictionaries include polynomial function bases, Fourier modes, spectral elements or other sets of (differentiable) functions of the full state observables. The choice of the specific dictionary to employ is a hyper-parameter inherent to EDMD, and is typically done empirically; see~\cite{williams2015EDMD} for a discussion on this topic.}
Then the Koopman operator can be approximated by minimizing the total residual between snapshots, that is % \textcolor{blue}{the difference between }
\begin{equation}\label{eq:argmin}
%    \begin{aligned}
        J %&=\frac{1}{2} \sum_{m=1}^{M}\left(r\left(x_{m},u_m\right)\right)^{2} \\
        =\frac{1}{2} \sum_{m=1}^{M}\left(\mathbf{\Psi}\left(x_{m+1}\right)-\mathbf{\Psi}\left(x_{m}\right) K\right)^{2} \enspace,
    %\end{aligned}
\end{equation}

Solving the least-square problem~\eqref{eq:argmin} with truncated singular value decomposition yields
    \begin{equation}\label{eq:estimation}
        K \triangleq \boldsymbol{G}^{\dagger} \boldsymbol{A}
    \end{equation}
where $^\dagger$ denotes the pseudoinverse, and $G,A$ are given by
\begin{equation}\label{eq:G&A}
    \begin{cases}
      \boldsymbol{G}=\frac{1}{M} \sum_{m=1}^{M}  \mathbf{\Psi}_m^{*} \mathbf{\Psi}_m\enspace,\\
      \boldsymbol{A}=\frac{1}{M} \sum_{m=1}^{M}  \mathbf{\Psi}_m^{*}  \mathbf{\Psi}_{m+1}\enspace,
    \end{cases}
\end{equation}
with $T$ and $^*$ denoting transpose and conjugate transpose operations, respectively. (For clarity we use $\mathbf{\Psi}_m$ to abbreviate $\mathbf{\Psi}(x_m,u_m)=[\psi_{1}(x_m,u_m), \ldots, \psi_{Q}(x_m,u_m)]$.)

Given $K$ computed via~\eqref{eq:estimation}--\eqref{eq:G&A}, the associated Koopman modes, eigenvalues and eigenfunctions are then computed as %($T$, $^*$ denotes the transpose and conjugate transpose.)
\begin{equation}\label{eq:KoopmanDecomposition}
    \begin{cases}
        v_q = (w_q^*B)^T \enspace,\\
        \lambda_q \xi_q = K\xi_q \enspace,\\
        \varphi_q = \mathbf{\Psi}_t\xi_q \enspace,
    \end{cases}
\end{equation}
where $\xi_q$ is the $q$-th eigenvector, $w_q$ is the $q$-th left eigenvector of $K$ scaled so $w_q^T\xi_q = 1$, and $B$ is the matrix of appropriate weighting vectors so that $x=(\mathbf{\Psi} B)^T$. By plugging expressions~\eqref{eq:KoopmanDecomposition} back to~\eqref{eq:extendPredicion} we can then describe the evolution of the system using the estimated Koopman operator. Note that $\mathbf{\Psi}_t=[\psi_{1}(x_t,u_t), \ldots, \psi_{Q}(x_t,u_t)]^T$, is the only term that needs (contains) information of current states. By using the Koopman operator and EDMD, all other terms can be computed from $M+1$ (training) measurements.

%\section{Problem Statement }\label{sec:problem}
%%
%%%%
%As described in last section, we could estimate the dynamic of a system through sensed data using Koopman operator theory. The estimated model could be later used for control, navigation, task assignment and other applications in robots. While when the measurements for training are noisy, the prediction will be inaccurate. So for an existing algorithm that using the Koopman operator for modeling as shown in Fig.~\ref{fig:system}, we will propose the prediction error in this paper to account for the effect of noise in training data, which in turn provides robustness guarantee for those methods using Koopman operator theory. 

%%%%%%%

\section{Quantification of Prediction Error}\label{sec:prediction}

%As described in last section, we could estimate the dynamic of a system through sensed data using 
The Koopman operator theory can thus be used to estimate the dynamics of a system directly from data (as described above). %The estimated model could be later used for control, navigation, task assignment and other applications in robots. 
However, when training measurements are noisy, the prediction will be inaccurate. %
%So for an existing algorithm that using the Koopman operator for modeling as shown in Fig.~\ref{fig:system}, we will propose the prediction error in this paper to account for the effect of noise in training data, which in turn provides robustness guarantee for those methods using Koopman operator theory. 
%
In this section, we present the main technical result of the paper which seeks to quantify the prediction error because of noisy measurements.  
%The diagram for analysis, which is consist with the procedure of modeling using the Koopman operator, is described in Fig.~\ref{fig:algorithm}.

We first investigate the sensitivity of the approximated Koopman operator $K$ to noisy training measurements, followed by the sensitivity of the  eigendecomposition to disturbed elements of $K$. Then, we combine them together to quantify the prediction error caused by noisy data when Koopman operator theory is employed. %We will list the procedure for obtaining disturbed term for the $i^{th}$ element here and other elements follow a similar process. 
%
%Note that we focus on the case when the Koopman operator is pretrained so the noise in current state $x_t$ is not considered in the approximation sensitivity.
%%%

\subsection{Sensitivity Analysis of Koopman Operator}
Perturbations on an element $x_m^i$ (i.e. the $i^{th}$ state of measurement at snapshot $m$ with $i\in\{1,\ldots,N_x\}$), will cause an error $\Delta K^i$. This can be obtained by chain rule on~\eqref{eq:estimation} and~\eqref{eq:G&A} as
\begin{equation}\label{eq:deltaK}
   \Delta K^i =[\Delta k_{ab}^i]_{Q\times Q}= \sum_{m=1}^{M+1}\{(\frac{\partial G^\dagger}{\partial x_m^i} A + G^\dagger \frac{\partial A}{\partial x_m^i})\Delta x_m^i\}\enspace, 
\end{equation}
where~\footnote{For details on the proof of the following pseudo-inverse terms see~\cite{golub1973differentiationPseudoinverse}.}
\begin{align*}
\frac{\partial G^\dagger}{\partial x_m^i} = -G^\dagger \frac{\partial G}{\partial x_m^i}G^\dagger &+G^{\dagger} (G^{\dagger })^*\left(\frac{\partial G^{*}}{\partial x_m^i} \right)\left(I-G G^{\dagger}\right)\\
&+\left(I-G^{\dagger} G\right)\left(\frac{\partial G^{ *}}{\partial x_m^i} \right)  (G^{\dagger})^*G^{\dagger}
\end{align*}

\begin{equation*}
        \frac{\partial A}{\partial x_m^i} =
        \begin{cases}   
        \frac{1}{M}\mathbf{\Psi}_{m+1}^{*}\frac{\partial\mathbf{\Psi}_m}{\partial x_m}\mathbf{e_i},\enspace \text{if } m=1 \\   
        \frac{1}{M}\mathbf{\Psi}_{m-1}^{*}\frac{\partial\mathbf{\Psi}_m}{\partial x_m}\mathbf{e_i},\enspace \text{if } m = M+1 \\ 
        \frac{1}{M}(\mathbf{\Psi}_{m+1}^{*}+\mathbf{\Psi}_{m-1}^{*})\frac{\partial\mathbf{\Psi}_m}{\partial x_m}\mathbf{e_i},\enspace \text{else}
        \end{cases}
\end{equation*}
and
\begin{equation*}
        \frac{\partial G}{\partial x_m^i} =
        \begin{cases}   
        0,\enspace \text{if } m = M+1 \\ 
        \frac{2}{M}\mathbf{\Psi}_m^{*}\frac{\partial\mathbf{\Psi}_m}{\partial x_m}\mathbf{e_i},\enspace \text{else} \enspace .
        \end{cases}
\end{equation*}
Here $\mathbf{e_i}$ represents the $i^{th}$ unit vector in $\mathbb{R}^{N_x}$.

\subsection{Sensitivity Analysis of Eigendecomposition}

As described in~\cite{crossley1969eigenvalue}, if a generic element $k_{ab}$ is perturbed, the eigenvalues and eigenvectors of the matrix $K=[k_{ab}]\in \mathbb{R}^{Q\times Q}$ are affected. The sensitivity of left eigenvectors $w_q$, right eigenvectors $\xi_q$, and eigenvalues $\lambda_q$ can be written as 
\begin{equation}\label{eq:eigenSensitivity}
    \begin{cases}
        c_{\lambda_q}^{ab}=\frac{\partial \lambda_q}{\partial k_{ab}} = w_q^a \xi_q^b,\enspace (a,b,q=1,2,\dots,Q)\enspace,\\
        c_{\xi_q}^{ab}=\frac{\partial \xi_{q}}{\partial k_{a b}}=%\zeta_{i}^{ab} \xi_{i}+
        \sum_{j=1 \atop j \neq q}^{Q} h_{j q} ^{ab} \xi_{j} \enspace,(a,b,q=1,2, \ldots, Q)\enspace,\\
        c_{w_q}^{ab}=\frac{\partial w_{q}^*}{\partial k_{a b}}=-%\zeta_{i}^{ab} w_{j}^T-
        \sum_{j=1 \atop j \neq q}^{Q} h_{q j} ^{ab} w_{j}^*\enspace,(a,b,q=1,2, \ldots, Q)\enspace,
    \end{cases}
\end{equation}
where $w_q^a$ denotes the $a^{th}$ element of the $q^{th}$ left-eigenvector $w_q$, $\xi_q^b$ is the $b^{th}$ element of the $q^{th}$ right-eigenvector $\xi_q$, and $h_{q j}^{ab}$ represents the $a^{th}$ row and $b^{th}$ column element of matrix $H_{qj}$ (similar for $h_{j q}^{ab}$). The matrix $H_{qj} = \left[h_{q j}^{ab}\right]_{Q \times Q}$ is computed by $\frac{w_{q} \xi_{j}^{*}}{\lambda_{j}-\lambda_{q}}$. 
That is, to get the eigendecomposition sensitivity of each element $k_{ab}$, we first need to obtain ($Q \times Q-Q$) $H$ matrices, and then every entry of these matrices will be used as parameters in~\eqref{eq:eigenSensitivity}.

\subsection{Sensitivity Analysis of Predicted Output}

We are now ready to derive the perturbation in predicted outputs as a result of measurement noise. Setting $[I_{N_x \times N_x}, O_{N_x \times N_u}]$ as $R$, where $I_{N_x \times N_x}$ denotes the ${N_x \times N_x}$ identity matrix and $O_{N_x \times N_u}$ denotes a ${N_x \times N_x}$ zero matrix. For the fully-observable forced nonlinear discrete time system~\eqref{eq:extendPredicion}, we can describe its deviation as
\begin{equation}\label{eq:ori_formulation}
    \Delta x_{t+1} = R \sum_{m=1}^{M+1}\left(\sum_{q=1}^Q\sum_{i=1}^{N_x}\frac{\partial(v_q\lambda_q\varphi_q(x_t,u_t))}{\partial x_m^i}\Delta x_m^i\right)
\end{equation}

$$ = R\sum_{m=1}^ {M+1}\left(\sum_{a=1}^Q \sum_{b=1}^Q \sum_{q=1}^Q\sum_{i=1}^{N_x}\frac{\partial(v_q\lambda_q\varphi_q(x_t,u_t))}{\partial k_{ab}^i}\frac{\partial k_{ab}^i}{\partial x_m^i}\Delta x_m^i \right)$$
\begin{equation*}
    \begin{medsize}
    =  R\sum_{a}^Q \sum_{b}^Q\left((\sum_{q}^Q\sum_{i}^{N_x}\frac{\partial(v_q\lambda_q\varphi_q(x_t,u_t))}{\partial k_{ab}^i})(\sum_{m}^{M+1}\frac{\partial k_{ab}^i}{\partial x_m^i}\Delta x_m^i )\right).
    \end{medsize}
\end{equation*}
Letting 
\begin{equation*}
\Delta k_{ab}^i = \sum_{m=1}^{M+1}\frac{\partial k_{ab}^i}{\partial x_m^i}\Delta x_m^i, 
\end{equation*}
we deduce 
\begin{equation}\label{eq:Sensitivity}
        \Delta x_{t+1} = R\sum_{a=1}^Q \sum_{b=1}^Q\left( \sum_{q=1}^Q\sum_{i=1}^{N_x}\frac{\partial(v_q\lambda_q\varphi_q(x_t,u_t))}{\partial k_{ab}}  \Delta k_{ab}^i\right)\enspace.
\end{equation}

Note that as illustrated in~\eqref{eq:eigenSensitivity}, for a dynamical system evolving by $K$, $\frac{\partial \lambda_q}{\partial k_{ab}}=c_{\lambda_q}^{ab}$, $\frac{\partial \xi_{q}}{\partial k_{a b}}=c_{\xi_q}^{ab}$, and $\frac{\partial w_{q}^T}{\partial k_{a b}}=c_{w_q}^{ab}$ are finite (bounded) constants. Then, for every $k_{ab}$ we have 
\begin{equation}\label{eq:first_term}
    \begin{medsize}
    \begin{aligned}
    \frac{\partial(\boldsymbol{v}_q\lambda_q\varphi_q(x_t,u_t))}{\partial k_{ab}} = \frac{\partial(\boldsymbol{v}_q)\lambda_q\varphi_q}{\partial k_{ab}}
+\frac{\boldsymbol{v}_q\partial(\lambda_q)\varphi_q}{\partial k_{ab}}
+\frac{\boldsymbol{v}_q\lambda_q\partial(\varphi_q)}{\partial k_{ab}}\\
        =((c_{w_q}^{ab})^*B)^T \lambda_j\mathbf{\Psi}_t\xi_q
        +(w_q^*B)^T c_{\lambda_q}^{ab} \mathbf{\Psi}_t\xi_q
        +(w_q^*B)^T \lambda_q\mathbf{\Psi}_t c_{\xi_q}^{ab}.
    \end{aligned}
    \end{medsize}
\end{equation}

%Finally for the $i^{th}$ element, 
Using matrix format of~\eqref{eq:first_term} and $\Delta K^i = [\Delta k_{ab}^i]_{Q\times Q}$ from~\eqref{eq:deltaK}, we can first compute the dot product of these two matrices and then obtain the prediction error as the sum of each element of the dot product, that is 
\begin{equation}\label{eq:pred_error}
        \begin{medsize}
    \begin{aligned}
    \Delta x_{t+1}
    &= R e^T \{\left[\sum_{q=1}^Q\frac{\partial(\boldsymbol{v}_q\lambda_q\varphi_q(x_t,u_t))}{\partial k_{ab}}\right]_{Q\times Q} \cdot (\sum_{i=1}^{N_x}\Delta K^i)\}e \enspace, \\
    \end{aligned}
        \end{medsize}
\end{equation}
where $e = [1, 1,\dots,1]^T$. 

\begin{figure}[t!]
\vspace{6pt}
\centering
\includegraphics[width = 0.48\textwidth]{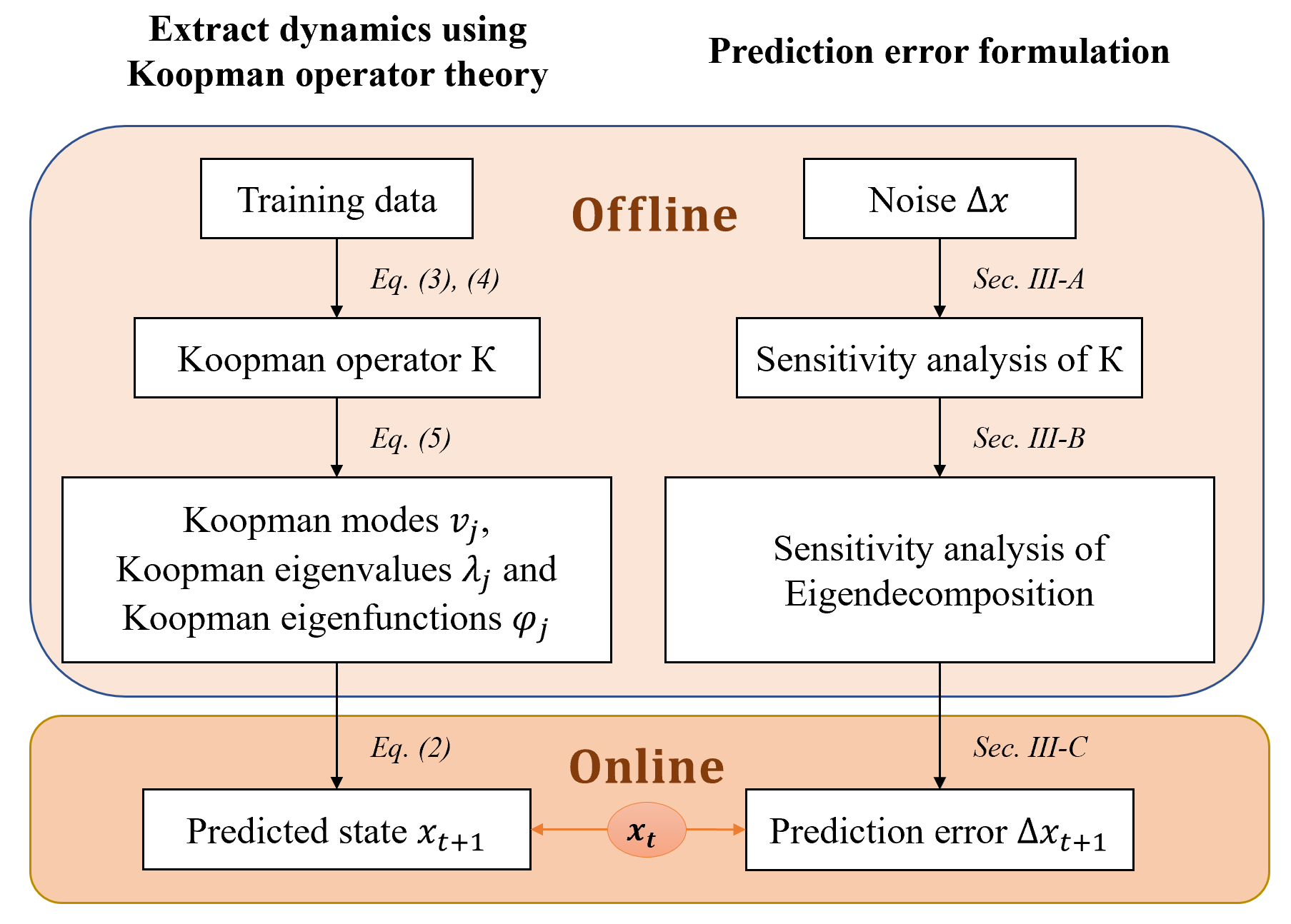}
\vspace{-18pt}
\caption{Illustration of the proposed approach's pipeline to adding robustness to data-driven robot control design (c.f. top system sub-block in Fig.~\ref{fig:system}). Our approach can be decomposed into an offline and an online component for real-time implementation, specific details of which are given in Algorithm~\ref{alg:alg}.}
\label{fig:algorithm}
\end{figure}

It is worth noting that in our proposed formulation, $\Delta K^i$ is estimated using only measurements and information about the noise, so it can be obtained offline. Similarly, the eigendecomposition sensitivity analysis of $K$ is predetermined and can yield the complex parameters $c_{w_q}^{ab}$, $c_{\lambda_q}^{ab}$, $c_{\xi_q}^{ab}$ ahead of time. These allow part of our approach to be executed during training, and enable the computation of the prediction error to take place at run-time, to adjust to measurements errors and thus endow robustness to the overall data-driven robot control architecture (Fig.~\ref{fig:system}). 
Finally the perturbed prediction is written as a function of the training measurements on states and inputs ($X$, $U$) of length $M+1$ % used to estimated the dynamics 
and current observation $\mathbf{\Psi}(\mathbf{x_t},\mathbf{u_t})$.  
For clarity of presentation, we will use the following abbreviation in the remainder of the paper 
\begin{equation}%\label{eq:prediction_error}
    \Delta \mathbf{x_{t+1}} = \eta (X,U, \mathbf{\Psi}(\mathbf{x_t},\mathbf{u_t}))\enspace.
\end{equation}
The relation between the offline (at training-time) and online (at run-time) components of our approach is shown in Fig.~\ref{fig:algorithm} and is discussed next.

%%%%
\section{Algorithmic Implementation}

%As formulated in last section, we could extract dynamics using Koopman operator and derive the prediction error as a function of the noisy training data together with current observations. 
The proposed quantification of prediction error approach describes analytically how to accommodate for measurement noise in data-driven control systems that are based on the Koopman operator. To make the approach amenable to robot control applications, where real-time implementation is required, we demonstrate in this section how the most computationally-intense aspects of the proposed approach can in fact be precomputed. Then, we present the algorithmic implementation of our approach in Algorithm~\ref{alg:alg}. 
%In this section, we discuss some practical considerations when we want to implement this data-driven method for robot in real-time. 
%A detailed algorithm is given and we will show how it works in the simulations of a Van del Pol oscillator system.  }

\subsection{Enabling Real-time Operation}
%
%\begin{enumerate}
%    \item
\textbf{Reformulation of dynamics-extraction expression:}
    Recall that only the observation at current time $t$, $\mathbf{\Psi}_t$, requires information gathered during run-time for making a prediction. We can thus rewrite~\eqref{eq:extendPredicion} to extract the terms that can be precomputed. Consider the dynamics
    $$
    x_{t+1} = [I_{N_x \times N_x}, O_{N_x \times N_u}]\sum_{q=1}^Q v_q\lambda_q\varphi_q(x_t, u_t)\enspace.
    $$
   Combining~\eqref{eq:KoopmanDecomposition} yields
\begin{align*}
    \sum_{q=1}^Q v_q\lambda_q\varphi_q(x_t, u_t)&=
    \sum_{q=1}^P (w_q^*B)^T \lambda_q \mathbf{\Psi}_t\xi_q\\
    &=\left(\sum_{q=1}^Q  \left(\mathbf{\Psi}_t\xi_q\right)^T \left((w_q^*B)^T \lambda_q\right)^T \right)^T\\
    %Note that $\left(\mathbf{\Psi}_t\xi_q\right)$ is a scalar (and hence $\left(\mathbf{\Psi}_t\xi_q\right)=\left(\mathbf{\Psi}_t\xi_q\right)^T$), its transpose is same as itself, so we could rewrite the equation as:
    %$$\sum_{q=1}^Q v_q\lambda_q\varphi_q(x_t, u_t)=
   &=\left(\sum_{q=1}^Q  \mathbf{\Psi}_t\xi_q\ \left((w_q^*B)^T \lambda_q\right)^T \right)^T
\end{align*}

    \begin{equation}\label{eq:F}
            \to x_{t+1}= [I_{N_x \times N_x}, O_{N_x \times N_u}] \left(\mathbf{\Psi}_t F \right)^T \enspace,
    \end{equation}
    where $F = \sum_{q=1}^Q \xi_q \lambda_q w_q^*B$. 
Thus, we have separated~\eqref{eq:extendPredicion} into offline ($F$) and online ($\mathbf{\Psi}_t$) parts.

    %%%%%%%
%%%%   

    % \item \textbf{Pseudo-inverse:} why pseudo-inverse instead of inverse : to ensure that the solution is unique, the optimization problem is solved by truncated singular value decomposition. If the determinant of the matrix is zero it will not have an inverse and your inv function will not work. This usually happens if your matrix is singular.
    
%    \item 
\textbf{Information of noise:}
    In the original formulation~\eqref{eq:ori_formulation}, we require the noise in measurements $\Delta x_m$ at every time instant. However, in practice we only have access to (or make assumptions about) noise statistics, for example mean $E(\Delta x)$ and standard deviation $\delta(\Delta x)$ in the case of Gaussian noise. 
    We can thus replace each $\Delta x_m$ by $E(\Delta x)$ directly. Also, we can randomly sample a training dataset $\Delta X \sim (E(\Delta x),\delta^2(\Delta x))$, applying it to~\eqref{eq:ori_formulation}, and repeat the process for several times to get the average values for accuracy.
    \footnote{The case of Gaussian noise herein is used as an example. Our proposed approach does not depend on the type of noise employed. Rather, noise statistics is a hyper-parameter that can be manually selected by the user or approximated via any available training data.}
    
%\end{enumerate}
%%%%

%\subsection{Algorithm}
%\kkm{We will summarize the whole procedure for (1) extracting dynamics using Koopman operator and (2) estimate the prediction error when we want to use the Koopman based data-driven methods for control. Each procedure will be clarified in "offline" and "online" phases as shown in Fig.~\ref{fig:algorithm}.}

\floatstyle{spaceruled}% Select new float style
\restylefloat{algorithm}% Apply spaceruled float style to algorithm
\begin{algorithm}[t]
\caption{Estimation Procedure}\label{alg:alg}
\textbf{Initialization:} Collect measurements of states $X$ and\\ inputs $U$ that might be noisy.
\\
\textbf{Offline training}:\\
\textit{Extracting dynamics}: 
\begin{itemize}
    \item Utilize $X$, $U$ to estimate the Koopman operator $K$ \\as well as matrices $G$, $A$ with~\eqref{eq:estimation} and~\eqref{eq:G&A}. 
    \item Compute parameter matrix $F$ in~\eqref{eq:F} with estimated $K$.
\end{itemize}
\textit{Estimation of prediction error}:
\begin{itemize}
    \item Obtain $\Delta K^i$ with information of noise in measurements and the approximated $K$, $G$, $A$ using~\eqref{eq:deltaK}.
    \item Get the parameters for eigendecomposition sensitivity $c_{\lambda_q}^{ab}$, $ c_{\xi_q}^{ab}$ and $c_{w_q}^{ab}$ through~\eqref{eq:eigenSensitivity} with estimated $K$.
\end{itemize}
\textbf{Online propagation}
\For{$t$}{
\Repeat{\text{control task is finished}}{
    \textit{Robotic System}: Use learned prediction error $\Delta x_t$ at last timestep to complement model constraints used in controller design. Then, drive the robot with this control signal.
    
    \textit{Extracting dynamics}:  
    \begin{itemize}
    \item Use $F$ to estimate the system dynamics and \\ do prediction as described in~\eqref{eq:F}.
    \end{itemize}
    
    \textit{Estimate prediction error}:
    \begin{itemize}
        \item Get eigen-sensitivity matrix~\eqref{eq:first_term} with \\pretrained\\ $c_{\lambda_q}^{ab}$, $ c_{\xi_q}^{ab}$, $c_{w_q}^{ab}$ and current observation $\mathbf{\Psi}_t$.
        \item Compute prediction error $\Delta x_{t+1}$ with $\Delta K$\\ and the eigen-sensitivity matrix by~\eqref{eq:pred_error}.
    \end{itemize}
\textbf{$t \gets t+1$}
}
}
\end{algorithm}
\vspace{-18pt}

%%%
%%%
\section{Parametric Testing and Evaluation}
\subsection{Simulation with a Van der Pol Oscillator}
To evaluate the proposed approach outlined in Algorithm~\ref{alg:alg}, we first test it using the Van del Pol oscillator~\cite{korda2018KoopmanMPC},
\begin{equation}\label{system:VandelPol}
\begin{cases}
\begin{array}{l}{\dot{x}_{1}=2 x_{2}} \\ {\dot{x}_{2}=-0.8 x_{1}+2 x_{2}-10 x_{1}^{2} x_{2}+u} \enspace.
\end{array}
\end{cases}
\end{equation}

The system~\eqref{system:VandelPol} is discretized with period $T_s = 0.01$\;s. $M+1$ pairs of data generated from the system are used to estimate Koopman operator $K$, and are further implemented as dynamic constraint in the controller. A random input vector is applied to propagate the system. 

\begin{figure}[h]
\vspace{-6pt}
\centering
\includegraphics[trim={0.15cm 0.15cm 1.75cm 0.50cm},clip,width = 0.38\textwidth]{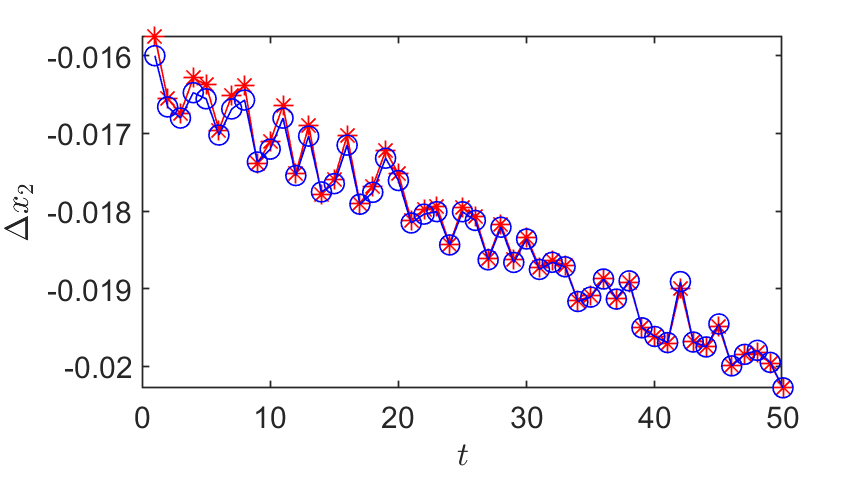}
\vspace{-6pt}
\caption{Prediction (blue circle) and true (red star) error under middle level noise ($c_2$) when $M=2*10^4$.}
\label{fig:middlenoise3}
\vspace{-6pt}
\end{figure}

The whole process is implemented in a parametric study as follows. We consider four distinct cases for measurement sets with 
$M = \{2*10^3,1*10^4,2*10^4,2*10^5\}$. We also consider three distinct noise levels for the disturbance. The amplitudes of the disturbances are chosen randomly from uniform distributions over the intervals $c_1 = [0,0.1x(0)]$, $c_2= [0,0.2x(0)]$ and $c_3=[0,0.4x(0)]$ for `\textbf{low} ($10\%$),' `\textbf{middle} ($20\%$)' and `\textbf{high} ($40\%$)' noise levels, respectively. 
%$c_3=40\%x(0)[0,1]$
These to lead to a total of $12$ distinct case studies.  For each case study, we train the system according to the selected measurements and noise settings, then give a random input vector to propagate the trained system for $50$ time steps, collect the output, and finally measure the true and predicted errors. 
To avoid bias and randomness, we evaluate the trained system under each case study in $5$ repeated trials, and averaged results are reported. For instance, Fig.~\ref{fig:middlenoise3} shows the average predicted error and the average true error when $M=2*10^4$ and under the $c_2$ noise case (middle).

%under varying measurement noise levels for $50$ time steps. The amplitudes of the disturbances are chosen randomly from uniform distributions over the intervals $c_1 = 10\%x(0)[0,1]$, $c_2= 20\%x(0)[0,1]$ and $c_3=40\%x(0)[0,1]$ for '\textbf{low}', '\textbf{middle}' and '\textbf{high}' density respectively, which lead to $12$ cases in total. 
%To avoid bias and randomness, each case is repeated $5$ times and the average curve is calculated. For instance, when $M=2*10^4$, the predicted and true error under the middle level noise are plotted in Fig~\ref{fig:middlenoise3}.
%

To compare overall the cases, we define the Mean Squared Error (MSE) between the \textbf{estimated} and \textbf{true} prediction error as
%To compare over all the cases, we define the absolute difference $D_{\mathrm{abs}}$ and relative difference $D_{\mathrm{r}}$ as:
\begin{multline}\label{eq:mse}
    MSE = \frac{1}{50}\sum_{t= 1}^{50}(\Delta x_2^{\mathrm{prediction}}(t)-(\Delta x_2^{\mathrm{true}}(t))^2\enspace.
\end{multline}
% \begin{multline}\label{eq:mse}
%     D_{\mathrm{abs}} = \frac{1}{50}\sum_{t= 1}^{50}|(\|\Delta x_2^{\mathrm{prediction}}(t)\|)-(\|\Delta x_2^{\mathrm{true}}(t)\|)|\enspace.
%     %D_{\mathrm{r}} = \sum_{t= 1}^{50}\frac{|(\|\Delta x_2^{\mathrm{prediction}}(t)\|)-(\|\Delta x_2^{\mathrm{true}}(t)\|)|}{\|\Delta x_2^{\mathrm{true}}(t)\|}\enspace.
% \end{multline}

\begin{figure}[h]
\vspace{-6pt}
\centering
\includegraphics[clip,width = 0.4\textwidth]{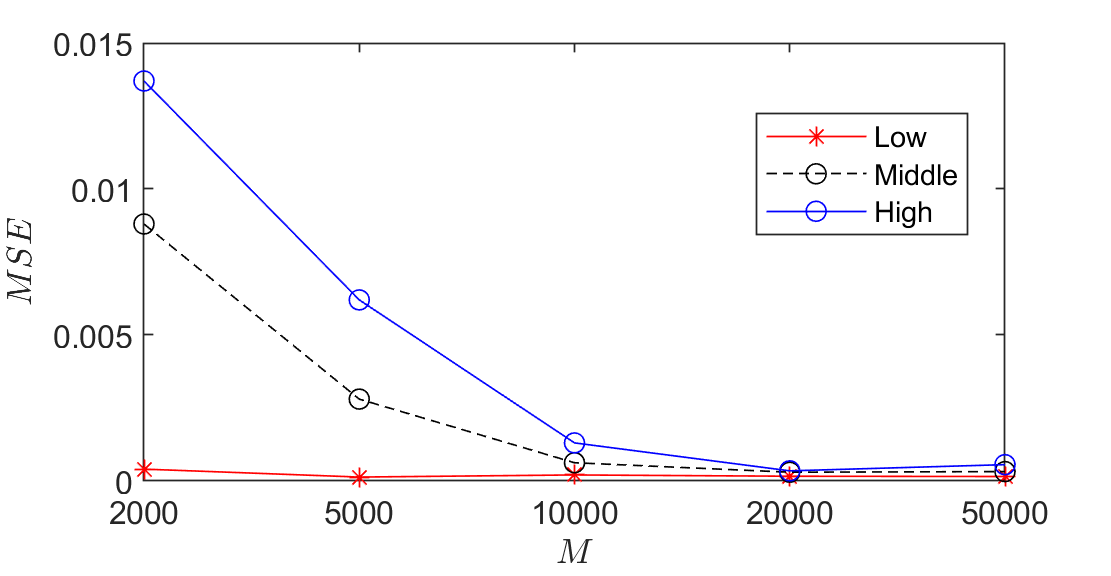}
\vspace{-12pt}
\caption{Mean Squared Error (MSE) of estimated and true prediction error under different noise levels. %(The approximately vertical line means the difference is too large to be plotted in the figure [when $M=200$].)
}
\label{fig:difference}
\vspace{-6pt}
\end{figure}

Results (averaged $MSE$) are depicted and compared in Fig~\ref{fig:difference}.
We observe that 1) low density noise leads to smaller prediction error for most choices of $M$. 2) Our method performs better with a large enough size of training data set (e.g., greater than $2*10^4$ as shown in the simulation).\footnote{In practice, good results may be achieved with fewer data, as we show next in simulation with a wheeled robot. Better understanding the interplay between training data set size and performance is part of ongoing work.} 3) If we keep increasing the size (e.g., $5*10^4$ in this one), its accuracy may not be improved further. %3) As the level of noise density decreases, we can better approximate the prediction error generated by noisy training data. 

% Results (the average of $D_{abs}$ and $D_r$) are depicted and compared in Fig~\ref{fig:difference}.
% We observe that 1) although low density noise leads to smaller absolute prediction error, it can cause a larger relative difference for most choices of $M$. 2) The method performs better with a large enough size of training data set (e.g., greater than $2*10^4$ as shown in that simulation). If we keep increasing the size (e.g., $2*10^5$ in this one), its accuracy may not be improved further. 3) As the level of noise density increases, we can better approximate the prediction error generated by noisy training data. 

%%%%
%%%%
\subsection{Simulation Experiments with a Wheeled Robot}
\begin{figure}[h]
\vspace{-6pt}
\centering
\includegraphics[clip,width = 0.16\textwidth]{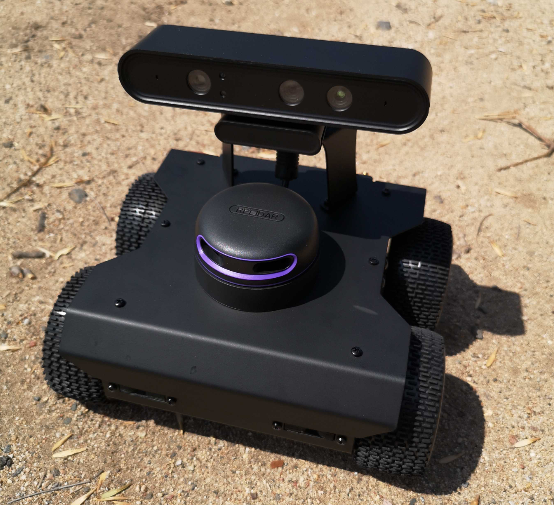}\includegraphics[clip,width = 0.168\textwidth]{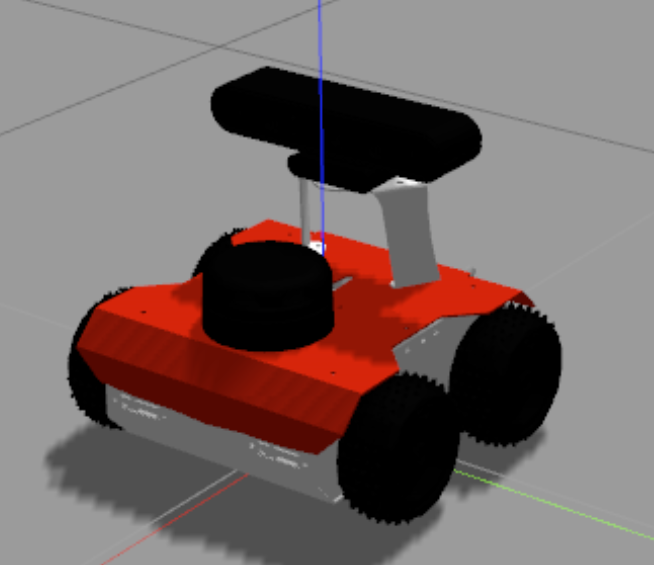}
\vspace{-6pt}
\caption{ROSbot 2.0 and its Gazebo model.}
\label{fig:rosbot}
\vspace{-6pt}
\end{figure}

Simulation experiments are conducted in the Gazebo environment using a ROSbot 2.0 robot, a differential-drive wheeled robot (Fig.~\ref{fig:rosbot}). The ROSbot 2.0 is an autonomous, open source robot platform and the embedded software allow us to send motion commands by manipulating the $x$ component of linear speed vector as $u_x$ and the $z$ component of the angular speed vector as $u_z$. The state vector contains the geometric center position $\{x,y\}$ and orientation $\theta$. %Its dynamics can be found in~\url{https://husarion.com}. The ROSbot 2.0 is an autonomous, open source robot platform and the embedded software allow us to send movement commends by manipulating the $x$ component of linear speed vector as $u_x$ and the $z$ component of the angular speed vector as $u_z$.

%%%%
%%%%

We test the efficiency of our proposed approach by building on top of an existing Koopman operator based controller~\cite{CCTA},\footnote{We use EDMD here to approximate the Koopman operator instead of DMD as in~\cite{CCTA}.} which is a data-driven hierarchical structure that refines the nominal input $u_o$ as $\hat{u} = u(K,u_o)$ to improve the performance of the controller under uncertainty. We highlight that the relation between inputs $u$ and the states $x$ is learned using the Koopman operator, and then the learned model is used to get a predicted (refined) input signal $\hat{u}$. Accurate and noisy data are used to generate the deterministic Koopman operator $K$ and stochastic one $K_n$. Then, these two are worked as model constraints to design the `\textbf{Nominal}' and `\textbf{Noisy}' controllers separately. To reduce the effect of noise in the learned model $K_n$, which is the contribution of our paper, we try to learn the prediction error generated by $K_n$ %compared to $K$
using Algorithm~\ref{alg:alg}. Then, this prediction error is used to complement the `Noisy' controller to yield the `\textbf{Proposed}' control signal. %Finally, the trajectories of the robot are generated by applying these three controllers, and are compared against a desired `\textbf{Reference}' trajectory. 

We compute the approximated Koopman operator using data captured when the robot is operating in open loop based on a random chattering linear velocity and angular velocity input signals. We perform $10$ such open-loop trials. The noisy training data set is created by adding disturbance in the position states $x$ and $y$. We sample the magnitude of noise for each state from the uniform distribution over the interval $c_1 = [0,0.25 \max(\xi_r)]$, $c_2 = [0,0.50 \max(\xi_r)]$, $c_3 = [0,0.75 \max(\xi_r)]$, $c_4 = [0,1.00 \max(\xi_r)]$.

\begin{figure}[h]
\vspace{-6pt}
\centering
\includegraphics[clip,width = 0.45\textwidth]{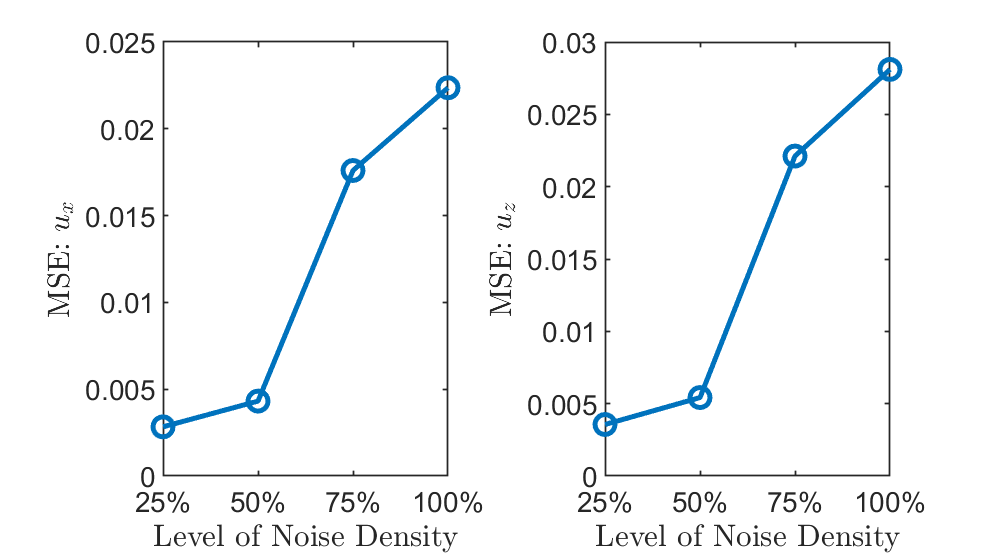}
\vspace{-6pt}
\caption{Mean Squared Error (MSE) of the estimated and true prediction error for linear velocity (left panel) and angular velocity (right panel) under different levels of noise.}
\label{fig:results}
\vspace{-6pt}
\end{figure}

\begin{figure}[h]
\vspace{-6pt}
\centering
\includegraphics[clip,width = 0.45\textwidth]{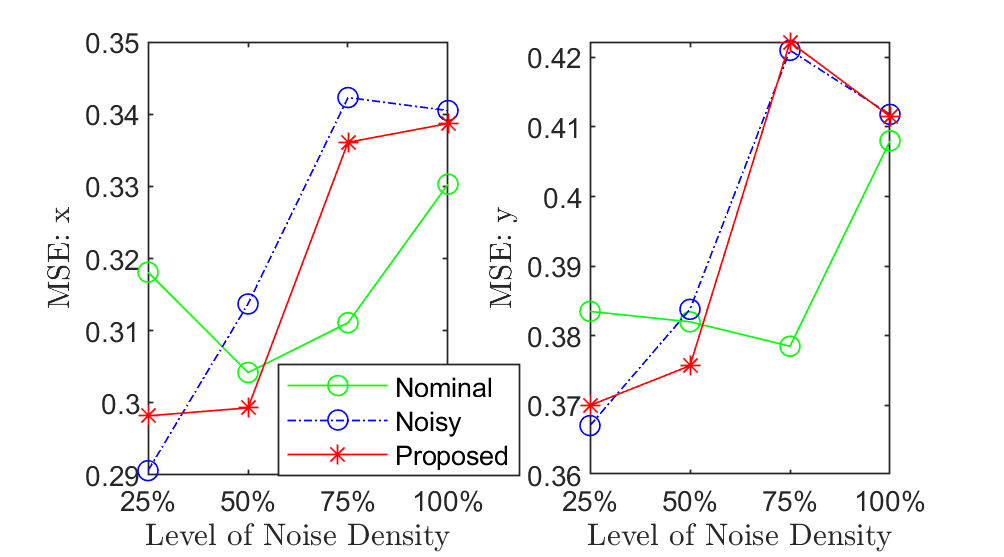}
\vspace{-6pt}
\caption{MSE of output trajectories' evolution along the $x$ axis (left panel) and $y$ axis (right panel) under distinct noise levels.}
\label{fig:results_x}
\vspace{-6pt}
\end{figure}

During testing, the robot is required to follow a semicircle trajectory for length $\frac{T}{2}$, i.e. $[x_r,y_r] = [\sin(\frac{2\pi}{T}t), 1-cos(\frac{2\pi}{T} t)]$.  %As mentioned, we will compare the performance of three controllers in the same environment. So 
To avoid bias, testing trials are repeated for $10$ consecutive times for an average with the same training set. We compute the Mean Squared Error (MSE) for the direct prediction error of $u_x$ and $u_z$ as defined in~\eqref{eq:mse}, where the \textbf{prediction} term is $\Delta u$ and the \textbf{true} term is $(u(K_n)-u(K)$ and the results %\footnote{As the magnitude of absolute value of error data under different level of noise is different, we use the true error to attain the relative one for authenticity.}
are illustrated in Fig.~\ref{fig:results}. We also obtain the difference among all trials for the output trajectories and show the results of position $x$ as well as $y$ in Fig.~\ref{fig:results_x}. 
%We compute the Mean Square Error (MSE) among all trials for the output trajectories (MSE$ = \frac{1}{n}\sum_1^N ([x, y]-[x_r,y_{r}])^2$) and for the direct prediction error (MSE$ = \frac{1}{n}\sum_1^N (\Delta u-(u(K_n)-u(K)))^2$).
%among all trials for the output trajectories with respect to the reference trajectory as defined in~\ref{eq:mse}}. and for the direct prediction error (MSE$ = \frac{1}{n}\sum_1^N (\Delta u-(u(K_n)-u(K)))^2$). Results\;\footnote{As the magnitude of absolute value of error data under different level of noise is different, we divide the MSE with the true error to attain the relative one for authenticity.} are illustrated in Fig.~\ref{fig:results} and Fig.~\ref{fig:results_x}. 
It can be observed that, 1) our proposed method can approximate the prediction uncertainty because of noise with a small error (Fig.~\ref{fig:results}). 2) When we implement the estimated uncertainty to a Koopman-based data-driven structure, our `\textbf{Proposed}' controller can drive the `\textbf{Noisy}' one closer to a desired trajectory. 3) As the noise level increases, our approach performs worse in terms of the prediction error estimation while performs better when we use the estimation for control. 4) Our algorithm remains bounded to the performance of the nominal controller (as the `$25\%$` noise level case in Fig.~\ref{fig:results_x}). Ongoing work focuses on implementation in different types of Koopman operator-based nominal controllers.

%3) Our approach performs better as noise levels increase (which is our target). 4) There appears to be a threshold on the noise level above which there may be no significant improvement (in the specific case considered here this happens as we reach the `$100\%$' noise level case). Further investigation along this line is part of future work.

% \begin{table}[h!]
% \vspace{-3pt}
% \caption{MSE and Standrad Deviation (STD) 25 noise}
% \label{tab:mse}
% \vspace{-12pt}
% \begin{center}
% \begin{tabular}{c|c|c|c}
% \hline
% Controller & MSE   & STD & MSE  \\
% &(w.r.t. Reference)& & (w.r.t. Original)\\
% \hline
% Nominal& 38.8137& [4.2699;3.5317]&0\\
% \hline
% Noisy&38.9326 & [4.8734;4.3649]&0.9234\\
% \hline
% Proposed &41.9677 & [5.2031;3.5236]& 0.7597\\
% \hline
% \end{tabular}
% \end{center}
% \vspace{-12pt}
% \end{table}

%%%%
%%%%
\section{Conclusions}

In this paper we focus on motion control of data-driven systems that are based on the Koopman operator theory to extract system models from data. We investigate the 
prediction error of a perturbed systems' performance when the data used for training the Koopman operator are noisy. The prediction error is then used to develop a new enhanced mobile robot motion control algorithm that endows robustness to data-driven systems. We show how certain aspects of our approach can happen offline, thus making the proposed algorithm amenable to real-time implementation. We test the proposed approach in a parametric simulated study using a Van der Pol oscillator to evaluate its performance as the number of training data and the level of noise corrupting them vary. We further test in Gazebo simulation with a non-holonomic wheeled robot tasked to track a reference trajectory. Results from both types of testing confirm that the proposed approach can apply across systems, including practical robotics. %, and that it performs better at low noise density levels.

%Then, we considered the practical implementation of the algorithm and do simplifications with respect to computation complexity. Taking care of some limitations of the approach when applying in the robot system, we propose the whole algorithm~\ref{alg:alg} so that one could easily follow the similar procedure and implement our work to enhance the robustness when they are using Koopman operator for modeling and control. Finally, we show the efficiency of our method in a Van del Pol oscillator and a differential drive robot ROSbot 2.0. The results imply that our algorithm can obtain the error of Koopman-based prediction under different levels of noise in training data. Also it performs better when there is higher uncertainty.
%
In all, our work provides robustness guarantees for systems using Koopman operator for control, and it can be viewed as a step toward developing new motion planners and controllers for data-driven mobile robots. Future research directions include integration with motion planning algorithms, and porting from Gazebo to hardware implementation. %On the other hand, we would also like to extend the idea to build the safety for other operator-based methods (e.g., Perron Frobenius transport operators).

%\newpage
\balance
\bibliographystyle{IEEEtran}
\bibliography{IEEEabrv, ICRA2021_koopman,PVSEM}

\end{document}